\documentclass[conference]{IEEEtran}
\IEEEoverridecommandlockouts
\usepackage{cite}
\usepackage{amsmath,amssymb,amsfonts}
\usepackage{algorithmic}
\usepackage{graphicx}
\usepackage{textcomp}
\usepackage{xcolor}
\usepackage{comment}
\usepackage{authblk}
\def\BibTeX{{\rm B\kern-.05em{\sc i\kern-.025em b}\kern-.08em
    T\kern-.1667em\lower.7ex\hbox{E}\kern-.125emX}}
\begin{document}

\title{Interpreting Deep Learning based Cerebral Palsy Prediction with Channel Attention\\
%{\footnotesize \textsuperscript{*}Note: Sub-titles are not captured in Xplore and
%should not be used}
\thanks{This project is supported in part by the Royal Society (Ref: IES\slash R2\slash 181024 and IES\slash R1\slash191147).

Emails: manli.zhu@northumbria.ac.uk, qianhumen2-c@my.cityu.edu.hk, e.ho@northumbria.ac.uk, howard@cityu.edu.hk, hubert.shum@durham.ac.uk

* Corresponding author
}
}
%\author{\IEEEauthorblockN{1\textsuperscript{st} Given Name Surname}
%\IEEEauthorblockA{\textit{dept. name of organization (of Aff.)} \\
%\textit{name of organization (of Aff.)}\\
%City, Country \\
%email address or ORCID}
%\and
%\IEEEauthorblockN{2\textsuperscript{nd} Given Name Surname}
%\IEEEauthorblockA{\textit{dept. name of organization (of Aff.)} \\
%\textit{name of organization (of Aff.)}\\
%City, Country \\
%email address or ORCID}
%}

\author[1]{Manli Zhu}
\author[2]{Qianhui Men}
\author[1]{Edmond S. L. Ho}
\author[2]{Howard Leung}
\author[3*]{Hubert P. H. Shum}
\affil[1]{\small Department of Computer and Information Sciences, Northumbria University, Newcastle upon Tyne, UK} %\authorcr Email: {\tt \{uid1, uid2\}@usc.edu}\vspace{1.5ex}}
\affil[2]{\small Department of Computer Science, City University of Hong Kong, Kowloon, Hong Kong} %\authorcr Email: {\tt uid3@jpl.nasa.gov} \vspace{-2ex}}
\affil[3]{\small Department of Computer Science, Durham University, Durham, UK} %\authorcr Email: {\tt uid3@jpl.nasa.gov}

\vspace{-2ex}

\maketitle

\begin{abstract}
Early prediction of cerebral palsy is essential as it leads to early treatment and monitoring. Deep learning has shown promising results in biomedical engineering thanks to its capacity of modelling complicated data with its non-linear architecture. However, due to their complex structure, deep learning models are generally not interpretable by humans, making it difficult for clinicians to rely on the findings. In this paper, we propose a channel attention module for deep learning models to predict cerebral palsy from infants' body movements, 
% recognize cerebral palsy body behaviors in infants, 
which highlights the key features (i.e. body joints) the model identifies as important, thereby indicating why certain diagnostic results are found. To highlight the capacity of the deep network in modelling input features, we utilize raw joint positions instead of hand-crafted features. We validate our system with a real-world infant movement dataset. Our proposed channel attention module enables the visualization of the vital joints to this disease that the network considers. Our system achieves 91.67\% accuracy, suppressing other state-of-the-art deep learning methods.
%We utilize raw joint position feature instead of hand-crafted features such as joint displacement and limb angle to verify the ability and robustness of our proposed deep learning model. 
\end{abstract}


\section{Introduction}
Cerebral palsy (CP) is one of the most common childhood neurodevelopmental disorders in the United States \cite{cp2017}. It affects people's movements, and has an impact on growth progression and life quality. It is estimated that between 3 to 10 out of 1,000 children develop some form of cerebral palsy depending on gestational age at birth \cite{data2003}. Therefore, early diagnosis is essential as it leads to taking the treatment as early as possible for preventing bigger health issues. However, a definitive diagnosis can be challenging for clinicians as years of practical experience and training is required.

Early diagnosis of CP has been investigated by the diagnostic tool General Movements Assessment (GMA), however, significant time and resource investment is needed to train an assessor. Automated GMA has been studied in some works. Stahl et al. \cite{autogma2012} applied optical flow and statistical pattern recognition on early video recordings of infants' spontaneous movements for later CP prediction. Wu et al. \cite{2021} extracted the infant body 2D key points from RGB images. From the depth information in RGB-D videos, they obtained the 3D movements of infants in the supine position, and extracted the limb angle features for infant CP prediction.

Machine learning and deep learning-based automated systems for CP diagnosis have been proposed. %Orlandi et al. \cite{mlpredection} applied machine learning methods such as Logistic Regression, AdaBoost, and Random Forest to detect atypical and typical infant movements using features extracted from videos.%, and 9 best features have been verified as the most important variables for the detection. 
McCay et al. \cite{Kevin2019} proposed the histograms of joint orientation 2D (HOJO2D) and joint displacement 2D (HOJD2D) features, for CP prediction by using machine learning methods such as K-nearest neighbors (KNN) and linear discriminant analysis (LDA).  %, and machine learning based methods are presented. Later in
They further proposed five deep learning models \cite{Kevin2020} to demonstrate the prospect of using deep learning in disease prediction. %using their proposed features. 
As manually crafted features
%the above mentioned methods heavily rely on feature selection, which %commonly consumes abundant effort and 
%involves a certain amount of human input as it 
%requires careful feature selection and 
require considerable domain expertise and may not be transferable to another domain, we opt for raw features, i.e. the joint positions, and demonstrate the same level of accuracy with our model.
%In this paper, we propose a deep learning framework, which is able to process the natural data in their raw form (joint positions) for CP diagnosis.

Although deep learning systems achieved impressive performance in disease diagnosis, it is hard for humans to understand how the decisions are made due to their black-box nature \cite{explainableAI2017}. A medical diagnostic system needs to be interpretable that allowing physicians and regulators to have the conﬁdence on the diagnostic result. 

To this end, we propose a channel attention module that tells what features have been considered important in our deep learning model during predictions. Attention mechanisms \cite{senet2018} have been widely used in deep learning for feature modelling, but there are limited researches on using them as an interface for interpreting deep neural networks. We implement our attention module with the squeeze-and-excitation framework \cite{senet2018}, which allows a single attention value to be learned for each channel. We propose a novel design to represent each joint as a channel, such that the attention value can be interpreted by humans.

We validate our system with an infant movement dataset \cite{dataset2018} targeting for CP prediction. We show that the inclusion of our module not only allows users to interpret the network at a feature level, but also achieves state-of-the-art performance comparing to existing deep learning methods that use hand-crafted features, demonstrating a reliable framework to analyze and predict CP on infants. 
%even though we do not use hand-crafted features.
Our source code can be downloaded at \textit{bit.ly/InterpretableDLforCP}.
%http://XXX.XXX/XXX.zip

%It does not adversely impact prediction accuracy and .
%joints are important of the movements for the diagnosis decision.  

\begin{figure*}[htbp]
	\setlength{\abovecaptionskip}{5pt} 
	\centering
	\includegraphics[width=0.92\linewidth]{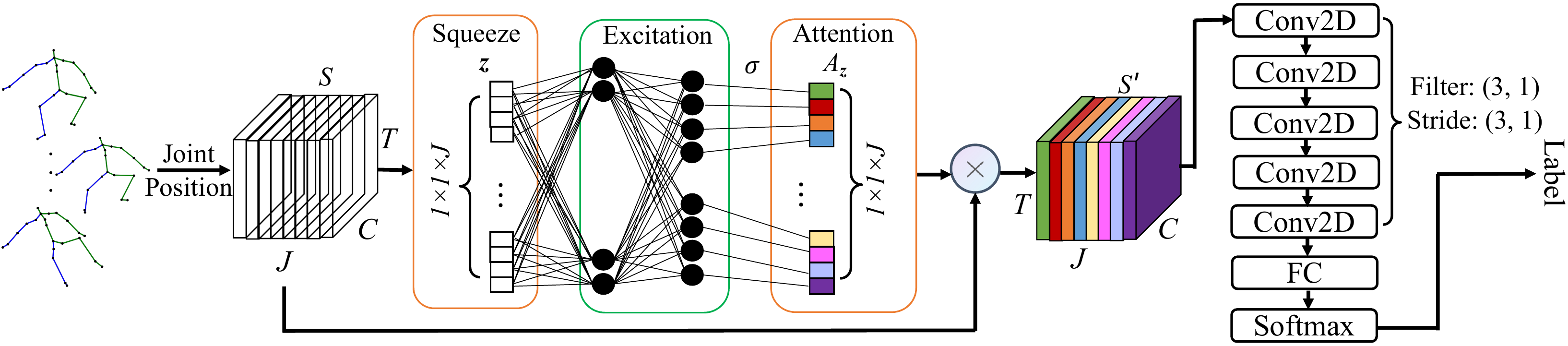}
	\caption{\footnotesize{The overview of our proposed framework.}}
	\label{Fig: framework}
\end{figure*}

\section{Methodology}
%[please add an overview of the framework and reference to Fig.1 here]
The overview of our proposed framework is illustrated in Fig. \ref{Fig: framework}. First, the input of the network is a temporal sequence of joint positions represented as a tensor (i.e. a multi-dimensional matrix). 
%we represent the input joint positions of a video sequence as an image in the form of a feature map, which is 
It is then fed into an attention module including Squeeze and Excitation blocks, and the importance of each joint is represented by its respective channel attention value. The attention value is multiplied with the input feature, forming an attended feature tensor. Such a tensor is fed into five 2D convolutional layers (Conv2D) to further extract the high-level information. Finally, a fully-connected layer (FC) and a softmax layer are applied to perform the classification.

\subsection{The Dataset}
We conduct experiments on the public dataset Moving Infants In RGB-D (MINI-RGBD) \cite{dataset2018}, in which twelve sequences of real infant movements were captured. To protect the privacy of recorded infants, such movements are represented by synthetic infant characters generated using the Skinned Multi-Infant Linear (SMIL) model \cite{smil2018}. In this research, we use the provided 3D joint positions of the sequences as our input. 

McCay et al. \cite{Kevin2019} further annotated the dataset, and each sequence was labeled as either {\it normal} or {\it abnormal} by an independent expert using the GMA method. We use such annotations as ground-truth.

%, in which twelve different sequences were generated using the Skinned Multi-Infant Linear (SMIL) model \cite{smil2018}. The MINI-RGBD dataset maps real infant movements to the synthetic SMIL 3D model to produce new textures and shapes, in order to protect the privacy of recorded infants. 

To ensure that the classifier is not affected by the global translation of the infant, we normalize the joint positions frame by frame. This is done by translating the spine joint of each frame to the global origin, and representing other joints as a relative value to the spine.

%In order to normalize the data and ensure that the classifier is not affected by the position, we take the spine as the center of the reference coordinate and calculate the difference between the first sequence and the other sequences of the spine position to relocate every infant body. 
%This re-positioning is carried out on all frames and on all sequences.

%\subsection{Data representation}
%Given a video sequence along with the 3D joint positions provided,
%The input 3D joint position features is denoted as a set $\mathcal{Q}=\left\{P_{t}^{j} \mid t=1,2, \ldots, T ; j=1,2, \ldots, J\right\}$, where $P_{t}^{j}$ denotes the joint $j$ at time $t$, $T$ denotes the total number of frames in the sequences, and $J$ denotes the total number of joints.
% of an infant full body within a frame.
	
%We first arrange the $X, Y$, and $Z$ 3D coordinates of joints in the length dimension, then we concatenate all the frames in the width dimension, and the sequence $S$ can be represented as a tensor with the following dimensions:
%\begin{equation}
%	\operatorname{dim}(\text {S})=\left(T, C, J\right)
%\end{equation}
%where $T$ is the size of length, $C$ denotes the number of width, $J$ is the number of channels.

\subsection{Channel Attention}
We propose an attention module based on the channel-wise attention mechanism \cite{senet2018} to provide insight into the network's decision process. As shown in our overall framework in Fig. \ref{Fig: framework}, we first propose a squeeze step that aims at compressing the joint information, then an excitation operation is applied to %the attention module includes squeeze and excitation two operations, the goal of it is to 
identify the importance of each joint to % which joints are the most important or less important for 
the network, such that different weights are assigned to different joints.

%We made a novel design to model the joints dimension as channels. This is because unlike traditional research to use channel attention for modelling features, we apply it in a novel manner for interpreting the network.
Unlike traditional research to use channel attention for modelling features, we model the joints dimension intuitively as channels as a novel way for interpreting the network. On the one hand, the joint dimension provides a more meaningful interpretation to the users compared to the coordinate dimension. On the other hand, 
%the designed attention module helps humans understand which joints contribute more to the diagnosis outcomes, which is more intuitive and efficient compared to learning the temporal attention added on the frames. 
using a single attention value for the whole 3D trajectory of a joint in the entire video results in a much more compact representation, %it has a smaller dimension comparing to the frame dimension, i.e. the result consists of fewer attention values, 
making it easier for humans to follow and more stable to train the network. As a result, the designed attention module helps humans understand which joints contribute more to the diagnostic outcomes.
%, and apply the designed attention module to help humans understand why a deep learning algorithm has come to a certain outcome of the diagnosis.

More specifically, the system input is represented as a feature tensor $S \in R^{T \times C \times J}$, %whose dimensions are $T \times C \times J$, 
in which $T$ is the total number of frames, $C=3$ represents the 3 dimensions of a joint position, $J$ is the total number of joints. In $S$, each feature value is represented as $S(t, c, j)$, where $t \in T$, $c \in C$ and $j \in J$. 

In the squeeze operation, the tensor from each joint is compressed into a representative joint descriptor using Global Average Pooling. 
%Given the whole feature map $T \times C \times J$, 
For a specific joint $j' \in J$, its descriptor $z_{j'}$ is calculated as:
\begin{equation}
z_{j'}=\frac{1}{T \times C} \sum_{t=1}^{T} \sum_{c=1}^{C} S(t, c, j')
%squeeze(S_{j})=
\end{equation}
These descriptors form a joint-level embedding $\mathbf{z}=\left[z_{1}, \ldots, z_{J}\right]$ from the whole movements. The obtained global embedding enables the aggregation of the local joint features which provides expressive joint-level statistics that triggers the following excitation steps.

The excitation operation calculates the attention value that indicates the importance of each channel (i.e. joint) by encoding the correlations from the embedded output $\mathbf{z}$ in the squeeze step. This is done by allocating the inner dependencies among joints with a nonlinear gating mechanism:
\begin{equation}
A_\mathbf{z}=\sigma\left(\mathbf{W}_{\mathbf{2}} \delta\left(\mathbf{W}_{\mathbf{1}} \mathbf{z}\right)\right)
\end{equation}
where $\delta$ is the ReLU activation, $\sigma$ is the sigmoid activation function, $\mathbf{W}_{\mathbf{1}}$ and $\mathbf{W}_{\mathbf{2}}$ are parameters of the fully-connected layers.
%, which are used for limiting model complexity and aiding generalisation.

The obtained $A_\mathbf{z} \in R^{1 \times 1 \times J}$ from the squeeze-and-excitation steps provides the joints attention, %with the dimension of $1 \times 1 \times J$, 
and each value in $A_\mathbf{z}$ indicates the attention weight of the corresponding joint. The attended feature tensor $S'$ is obtained by multiplying $A_\mathbf{z}$ with the original feature tensor $S$.  

\subsection{The CP Prediction Network}
We propose a 2D convolutional neural network to encode the tensor $S'$ for CP prediction. The network consists of five convolutional layers, one fully-connected layer and one softmax layer, as shown in Fig. \ref{Fig: framework}. We found that having multiple convolutional layers help to encode high-level features with a suitable receptive field, 
%gradually reduce the feature size 
and achieve the best classification accuracy. The fully-connected layer maps the learned feature with the CP labels (i.e. normal or abnormal ), and the Softmax layer is to generate probability distributions for classification.

The filter $K$ in the convolutional layers has the dimension of $F_{T} \times F_{C} \times J$, where $F_{T}=3$ and $F_{C}=1$ in our experiment. This means that the filter covers all joints at once, and moves in the frame and coordinate dimensions. This design ensures that the intrinsic correlations among joints can be discovered, and that the temporal correlations of nearby frames are modelled.
%The convolutional operation is defined as:
%\begin{equation}
%	\operatorname{conv2D}(S, K)_{m, n}=\sum_{t=1}^{F_{T}} \sum_{c=1}^{F_{C}} \sum_{j=1}^{J} K_{t, c, j} S_{m+t-1, n+c-1, j}
%\end{equation}
%where $m \in T$ and $n \in C$ are the row and column indices of the feature map $T \times C$.

%Formally, we define the filter with the following dimension:
%\begin{equation}
%	\operatorname{dim}(\text {K})=\left(F_{T}, F_{C}, J\right)
%\end{equation}
%where $F_{T}$ is the kernel size of width, $F_{C}$ denotes the kernel size of length. 
%In our experiment, they are set as 3 and 1, respectively for all convolutional layers.
%where $x$ and $y$ are the row and column indices, respectively.

%As shown in Fig. \ref{Fig: framework}, we apply five convolutional layers to extract the information from the weighted feature map, Then, feed the learned features into a fully connected layer.

\subsection{Loss Functions}
We propose a novel loss function known as attention loss. Unlike the traditional usage of attention, we apply attention for deep neural network interpretation. To facilitate better human understanding, the number of attended features should be minimized, while keeping the classification accuracy unchanged. We define the attention loss as:
\begin{equation}
L_{\text {att}} = \sum_{j=1}^{J} A_{z_{j}}
\end{equation}

The final loss function consists of three parts: the cross-entropy loss $L_{\text {cep}}$ to encourage classification accuracy, the attention loss $L_{\text {att}}$ to minimize the number of attended features, and the regularization loss $\|w\|_{2}$ (where $w$ is the network parameters) to discourage overfitting:
\begin{equation}
L=L_{\text {cep}}+\gamma L_{\text {att}} + \lambda\|w\|_{2}
\end{equation}
where $\gamma$ and $\lambda$ are the loss weights set as 0.0005 and 0.0001 respectively. We deliberately chose a small $\gamma$ such to minimize the attention loss effect to the classification accuracy.

%The regularization term is used for avoiding overfitting as our dataset is very small, and the attention loss ensures that the model is optimized by the attention weights as well. 

%The loss function is defined as:
%\begin{equation}
%L=L_{\text {cep }}+\lambda\|w\|_{2}+\gamma \sum_{j=1}^{J} A_{z_{j}}
%\end{equation}
%where the first term $L_{cep}$ is the cross-entropy loss function, the second term is the $L2$ regularization, and the third term is the attention loss. $\lambda$ and $\gamma$ are the loss weights set as 0.01 and 0.0001 respectively in our experiment.

%Lcep + L2+ 0.0001attention weights

%A paragraph on the loss functions: Cross entropy + L2 regularization
%Equation like $L_cross + \alpha L_regu$
%where $\alpha$ is the loss weight set as ...
%add one sentence for each loos to explain what they are

%Topic sentence: we use a class balancing loss as the data is biased.
%First, talk about class balancing loss, and how important they are. 
%Second, talk about how you implement it as below.

As the data is biased with fewer abnormal samples (i.e. 4 abnormal, 8 normal), we introduce a class balancing weight in $L_{cep}$. The idea is to give a higher weight to the abnormal samples to facilitate training. Without the weights, the model would have poor predictive performance for the minority class (i.e. abnormal) when an imbalanced dataset is given, resulting in more false-negatives.
In particular, we assign a weight $\alpha_{i}$ to the class $i$ by the following equation:
\begin{equation}
\alpha_{i}=\sqrt{\frac{n}{n_c \times n_{i}}}
\end{equation}
where $n$ is the total number of the samples, $n_{i}$ is the number of samples of class $i$, and $n_c = 2$ is the number of classes.

%such that the bias has minimal effect to the per-class classification accuracy - in this case, the false-negative can be reduced.

%as the data is biased. The number of normal samples is twice as much as abnormal samples in our used dataset. 
%Most of the machine learning algorithms are designed under the assumption of an equal number of samples for each class, and this results in the models have poor predictive performance for the minority class when an imbalanced dataset is given.

In our implementation, the whole network was trained in an end-to-end manner using the PyTorch platform with the Adams optimizer. The hyper-parameters \textit{epoch}, \textit{learning rate}, and \textit{batch size} were set as 400, 0.0003 and 3 respectively for all cross-validations. %Here , we use Adams optimizer for motion optimizing the model. 

\section{Experimental Results}

\subsection{Quantitative Evaluation}
%There are twelve sequences in the MINI-RGBD dataset, 
We perform the leave-one-out cross-validation on the MINI-RGBD dataset \cite{dataset2018}, which is also used in the baseline works \cite{Kevin2020, 2021}, and the averaged result for all cross-validations is presented as the ﬁnal accuracy.

As deep learning has become the mainstream in pattern recognition due to its capacity to modelling complicated data, we compare the classification performance of our proposed deep learning method with the existing deep learning methods \cite{Kevin2020} as well as the most recent GMA-based method \cite{2021} that is also based on 3D skeletal data as input. The results are reported in Table \ref{accuracy}.
%We use the same leave-one-out cross-validation evaluation and dataset as \cite{Kevin2020,2021}, so we simply copy their classification accuracy, and the results are reported in Table \ref{accuracy}.
It can be seen that our full system (i.e. the bottom row) achieves the same state-of-the-art performance, i.e. 91.67\% prediction accuracy, in addition to the interpretability of our network which tells users which of the joints of an infant body are the most important for its diagnostic outcome. Compared with \cite{Kevin2020}, we utilize raw features instead of hand-crafted ones (i.e. HOJO2D + HOJD2D that generates the best accuracy), which makes our method easy to be transferred to another domain as there is no feature engineering involved. Compared with \cite{2021}, which does not require any training for the model but a threshold has to be defined manually to separate the {\it positive} and {\it negative} samples. In contrast, 
%a huge amount of human analysis to carry out the diagnosis result, 
our model is trained end-to-end without the need for any human intervention. %, such that it provides the diagnosis result.% efficiently, effectively, and quickly. %Compared with \cite{2021}, in which a huge amount of human analysis is required, our model is trained end-to-end, we do not require considerable human inputs and experienced experts to carry out the diagnosis result as there is no feature engineering involved.

%on the joint position feature with all the given 23 joints instead of the manually extracted 14 joints used in \cite{Kevin2020,2021}, and this makes our method easily to be transferred to another domain. Especially, it outperforms the same deep learning based methods Conv1D-1, Conv1D-2, Conv2D-1, and Conv2D-2, in which the single type of feature HOJO2D or HOJD2D is used. By combining the two kinds of features, the performance of Conv1D-1 and Conv2-D also reaches 91.67\%, this indicating more features would contribute to improving the classification performance. 

%The latest GMA-based method also achieves 91.67\% accuracy using the limb angle feature, however, this method is supervised and it needs quite significant human inputs and experienced experts to carry out the diagnosis result. While our method is unsupervised, there is no feature engineering involved, and we believe it would have great potential in assisting clinicians with infant CP prediction.

\begin{table}[htbp]
	\centering
	\caption{Classification accuracy comparison between our proposed method and other baseline methods.}
	\label{accuracy}
	\setlength{\tabcolsep}{8pt}
	\begin{tabular}{|c|c|c|}
		\hline
		Feature & Method & Accuracy (\%)\\
		\hline
		HOJO2D & FCNet \cite{Kevin2020}& 83.33 \\
		HOJD2D & FCNet \cite{Kevin2020}& \textbf{91.67} \\
		HOJO2D / HOJD2D & Conv1D-1 \cite{Kevin2020}& 83.33 \\
		HOJO2D / HOJD2D & Conv1D-2 \cite{Kevin2020}& 83.33 \\
		HOJO2D / HOJD2D & Conv2D-1 \cite{Kevin2020}& 83.33 \\
		HOJO2D / HOJD2D & Conv2D-2 \cite{Kevin2020}& 83.33 \\
		HOJO2D + HOJD2D & Conv1D-1 \cite{Kevin2020}& \textbf{91.67} \\
		HOJO2D + HOJD2D & Conv1D-2 \cite{Kevin2020}& \textbf{91.67} \\
		Limb Angle & GMA-based method \cite{2021}& \textbf{91.67} \\ \hline 
		Joint Position & Ours - Without Attention & \textbf{91.67} \\
		Joint Position & Ours - Full System & \textbf{91.67} \\
		\hline
	\end{tabular}
\end{table}

We conducted an ablation study to evaluate if the attention module, which was responsible for providing the interpretation functionality, had any negative effect on classification accuracy. The results are shown at the bottom two rows of Table \ref{accuracy}. The attention module does not affect the overall classification accuracy, and it also provides us with an interface to interpret the outcome of the deep learning model in a human-understandable way. Notice that while traditional attention modules tend to improve classification accuracy, the target of ours is for interpretation. We tailor the way a channel is defined such that the results can be as human-understandable as possible. This limits the module's potential to fully optimize the attention for accuracy improvement. 
%We believe that there are two reasons why our attention module does not improve the classification accuracy. On the one hand, the dataset is too small, this limits the performance of our attention module. On the other hand, the target of designing the joints as channels is visualization for interpreting the deep learning model, therefore, it is not optimal for accuracy improvement. To further improve the accuracy, other designs may be considered. 
% Discussion: why our attention does not improve classification accuracy

%Put in an table of with and without the attention module, saying that our attention module does not harm classificaiton accuracy, but gives more information about interpretation.

%Put in an experiment of with and without the attention loss term, use quantitative results (e.g. average value of attention, number of attended joints, etc.) to say that the attention loss is working, also add in the classification accuracy to show that that loss term does not harm the classification accuracy.

\begin{table}[htbp]
	\centering
	\caption{The influence of the attention loss in our framework.}
	%\caption{Ablation study on the attention loss term.}
	\label{lossterm}
	\begin{tabular}{|c|c|c|}
		\hline
		%Method & Average Attention Value & \#over 0.9 & Accuracy (\%) 
		Method & Without $L_{att}$ & With $L_{att}$ \\\hline
		Avg. Per-joint Attention Value & 0.492 & \textbf{0.439} \\\hline
		Avg. \# of Joints with Attention $\geq$ 0.5 & 11.039 & \textbf{9.910} \\\hline
		Avg. \# of Joints with Attention $\geq$ 0.6 & 10.856 & \textbf{9.861} \\\hline
		Avg. \# of Joints with Attention $\geq$ 0.7 & 10.750 & \textbf{9.702} \\\hline
		Avg. \# of Joints with Attention $\geq$ 0.8 & 10.566 & \textbf{9.583} \\\hline
		Avg. \# of Joints with Attention $\geq$ 0.9 & 10.074 & \textbf{9.084} \\\hline
		Avg. \# of Joints with Attention $=$ 1.0 & 6.583 & \textbf{6.250} \\\hline
		Accuracy (\%) & \textbf{91.67} & \textbf{91.67} \\\hline
	\end{tabular}
\end{table}

We further studied the effectiveness of the proposed attention loss (see %, as shown in 
Table \ref{lossterm}). The present numbers are averaged by the total number of cross-validations. We find that the loss term effectively reduced the average per-joint attention values. The average number of joints with high attention values is also reduced. 
This allows the results to be more easily understood by humans, while maintaining the same prediction accuracy. 
	
\subsection{Quantitative Evaluation} % by Ed
\begin{figure}[htbp]
	\setlength{\abovecaptionskip}{0.cm} 
	\centering
	\includegraphics[width=.95\linewidth]{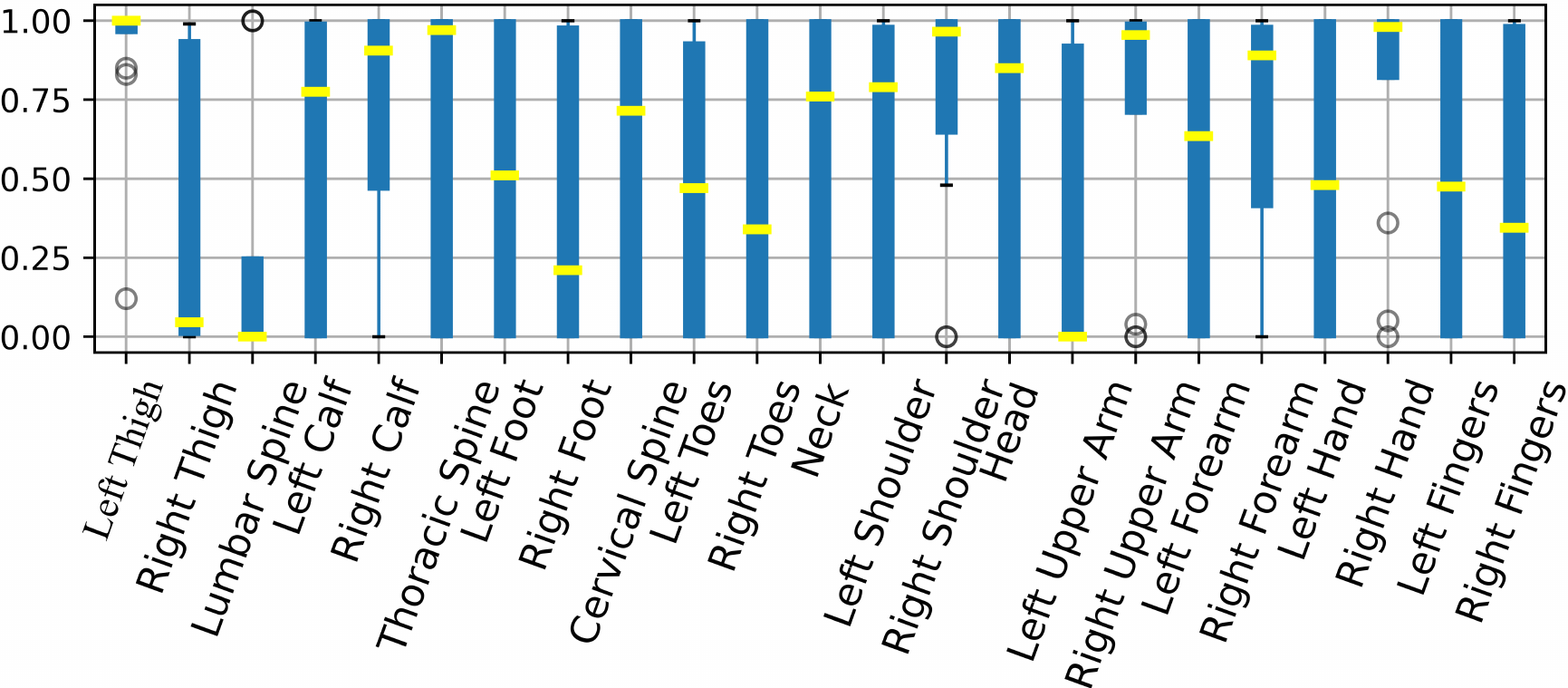}
	\caption{\footnotesize{The visualization of attention values of all joints across twelve validations.}}
	\label{Fig: attention}
\end{figure}

Fig. \ref{Fig: attention} presents a box plot to visualize the distributions of joint attention values across the twelve cross-validations. For each joint, the interquartile range (IQR) is represented by the blue bar, and the yellow line denotes the median value. The generally high median lines indicate most of the cross-validations tend to assign a higher attention value to the joint.

From the median lines, we can see that the joints `Left Thigh', `Right Hand', `Thoracic Spine', `Right Shoulder' and `Left Upper Arm' have the highest attention values in almost all the cross-validations. 
% By Ed - see if they are ok
Most of these joints are located in the upper body, including those on the arms and shoulders. This aligns with the results reported in \cite{Kevin2019} that the features extracted from the arms are more discriminative and resulted in higher accuracy in CP prediction on the MINI-RGBD dataset.
%These joints are `Left Thigh', `Lumbar Spine', `Right Foot', `Left Shoulder', `Right Hand', `Left Fingers', and `Right Fingers'. `Left Foot', `Cervical Spine', `Right Toes', `Head', `Left Upper Arm', and `Left Forearm' are also important for most tests. Indicating they are essential for the network to make the diagnosis decision.

\begin{figure}[htbp]
	\setlength{\abovecaptionskip}{0.cm} 
	\centering
	\includegraphics[width=0.95\linewidth]{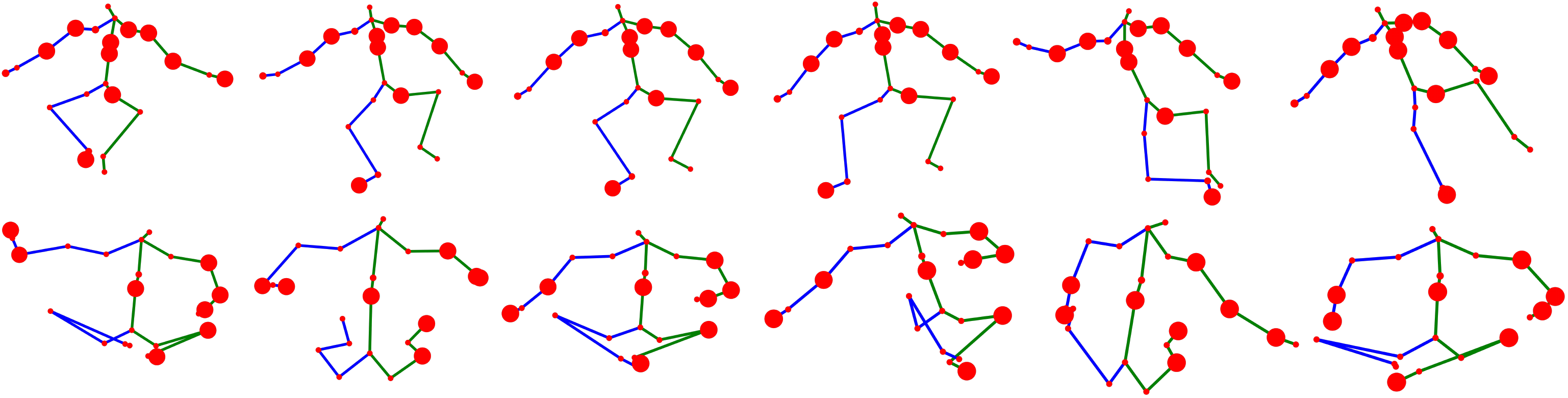}
	%\\ \hspace*{\fill} \\
	%\includegraphics[width=0.95\linewidth]{abnormal.pdf}
	\caption{\footnotesize{The visualization of joints attention in two cross-validations with a normal sample (upper row) and an abnormal sample (lower row) as validation.}}
	\label{Fig: visualization}
\end{figure}

We also visualize the joint attention values in two cross-validations, one with a normal sample (Fig. \ref{Fig: visualization} upper) and one with an abnormal sample  (Fig. \ref{Fig: visualization} lower) as validation. The sizes of the red circles visualize the attention values of the corresponding joints. It can be seen that the example motions from different classes (i.e. {\it normal} and {\it abnormal}) resulted in different sets of attention values. This highlights the network has different focuses on the body joints to differentiate motions from the two classes.
%The attention values are different in Fig. \ref{Fig: visualization} because the leave-one-subject-out cross-validation is used, and this results in slightly different sets of attention values. Notice that fewer joints were detected as important for the abnormal infant, who might demonstrate slow or few movements.
%The normal infant shows more attended joints than the abnormal one, this is because the abnormal infant may demonstrate slow or few movements, thus fewer joints are detected as important.
%For the selected two samples, most of the important joints are located in the upper body. 
Compared with the previous works \cite{Kevin2020, 2021}, in which manually select some joints for analysis, our attention module is able to identify the more important joints automatically.
%a part of two infants' movements in the form of the skeleton with highlighted joints, one is normal (Fig. \ref{Fig: visualization} upper) and the other one is abnormal (Fig. \ref{Fig: visualization} lower). 

\section{Conclusion}
We have presented an interpretable deep learning framework with channel attention for CP prediction. The attention module designed in this paper has shown how the network analyzes the movements by looking at the body joints that are more important to the final diagnostic results. We further demonstrated the effectiveness of the proposed attention loss for minimizing the number of attended joints such that the user can focus on a smaller set of joints for more in-depth analysis. Compared with the state-of-the-art methods, our proposed model achieves the same level of classification accuracy, even though we do not use hand-crafted features nor any manual interventions. 

% Discussion: future direction of have both spatial and temporal attention, a key challenge will be to represent the large number of temporal attention value in a meaningful manner

A future direction is to incorporate both spatial and temporal attention into our channel attention module, which will allow interpreting when an infant performs abnormal movements with which joints. 
This can facilitate a user to focus the analysis on the video segment with high attention.
A key challenge is the large number of frames, which results in a large number of temporal attention values. Further research on how to effectively visualize such information is needed.

%In the future, we will consider integrating temporal attention into the framework to identify the most important timestamps of an infant's movement. This provides the user with spatio-temporal information on when an infant performs abnormal movements,
%such that the user can further analyze those important video segments instead of going through the entire clip. 
%A key challenge of modelling temporal attention will be to represent a large number of temporal attention values in a meaningful manner and how to visualize such kind of spatio-temporal information to the users.

\bibliographystyle{IEEEtran}  
\bibliography{IEEEabrv,reference}

\end{document}